\pdfoutput=1

\documentclass[11pt]{article}

\usepackage[final]{acl}

\usepackage{times}
\usepackage{latexsym}
\usepackage{todonotes}
\usepackage[T1]{fontenc}

\usepackage[utf8]{inputenc}

\usepackage{microtype}

\usepackage{inconsolata}

\usepackage{graphicx}

\usepackage{booktabs} 
\usepackage{adjustbox}
\usepackage{todonotes}
\usepackage{amsmath}
\usepackage{amssymb}
\usepackage{multirow}
\usepackage{fontawesome} 

\usepackage{tcolorbox}
\usepackage{caption}

\usepackage{tabularx}
\usepackage{placeins}
\usepackage{booktabs}

\usepackage{enumitem}

\newtcolorbox{promptbox}{
 colback=white, 
 colframe=gray, 
 coltext=darkgray, 
 fonttitle=\bfseries, 
 sharp corners=downhill, 
 rounded corners, 
 boxrule=0.5mm, 
 width=\linewidth, 
 left=0.0em, 
 right=0.0em, 
 top=0.0em, 
 bottom=0.0em 
}

\newcommand{\prompt}[1]{%
 \begin{promptbox}%
 \scriptsize \texttt{#1}%
 \end{promptbox}%
}

\newcommand{\thumb}[2]{
 \textsc{#1}\textsuperscript{
 \tiny
 \ifthenelse{\equal{#2}{up}}{\faThumbsOUp}{\faThumbsODown}}%
}

\title{CrisiText: A dataset of warning messages for LLM training in emergency communication}

\author{
 \textbf{G. Gonella\textsuperscript{1, 2}},
 \textbf{G. M. Campedelli\textsuperscript{1}},
 \textbf{S. Menini\textsuperscript{1}},
 \textbf{M. Guerini\textsuperscript{1}}
\\
\textsuperscript{1}Fondazione Bruno Kessler, Italy,
\\
\textsuperscript{2}University of Trento, Italy
\\
 \texttt{\{ggonella, gcampedelli, menini, guerini\}@fbk.eu}
}

\begin{document}
\maketitle

\begin{abstract}
Effectively identifying threats and mitigating their potential damage during crisis situations, such as natural disasters or violent attacks, is paramount for safeguarding endangered individuals. To tackle these challenges, AI has been used in assisting humans in emergency situations. Still, the use of NLP techniques remains limited, and mostly focuses on classification tasks. The significant potential of timely warning message generation using NLG architectures, however, has been largely overlooked. In this paper we present \textit{CrisiText},
\let\thefootnote\relax\footnotetext{The \textit{CrisiText} dataset can be found at \url{https://huggingface.co/datasets/LanD-FBK/crisitext}}\let\thefootnote\svthefootnote 
the first large-scale dataset for the generation of warning messages across 13 different types of crisis scenarios. The dataset contains more than 400,000 warning messages (spanning almost 18,000 crisis situations) aimed at assisting civilians during and after such events. To generate the dataset, we started from existing crisis descriptions, and created chains of events related to the scenarios. Each event was then paired with a warning message. The generations follow expert’s written guidelines to ensure correct terminology and factuality of their suggestions. Additionally, each message is accompanied by three suboptimal warning types to allow for the study of different NLG approaches. To this end, we conducted a series of experiments comparing supervised fine-tuning setups with preference alignment, zero-shot, and few-shot approaches. We further assessed model performance in out-of-distribution scenarios and evaluated the effectiveness of an automatic post-editor.
\end{abstract}

\section{Introduction}

Our world is shaped by rapidly evolving social and environmental phenomena that can profoundly impact thousands or even millions. Terrorism and natural disasters exemplify the increasingly frequent risks that threaten entire communities, regions, and even nations \cite{lafree2007globalterrorism, WMO2021}. Consequently, there has been a growing interest in crisis and emergency management, aiming to bridge disciplines and develop practical, actionable solutions to mitigate risks from events such as tornadoes, floods, earthquakes, and violent acts like terrorism \cite{rosenthal2001managing, canton2019emergency}. 

\begin{figure}[t]
 \centering
 \includegraphics[width=1\linewidth]{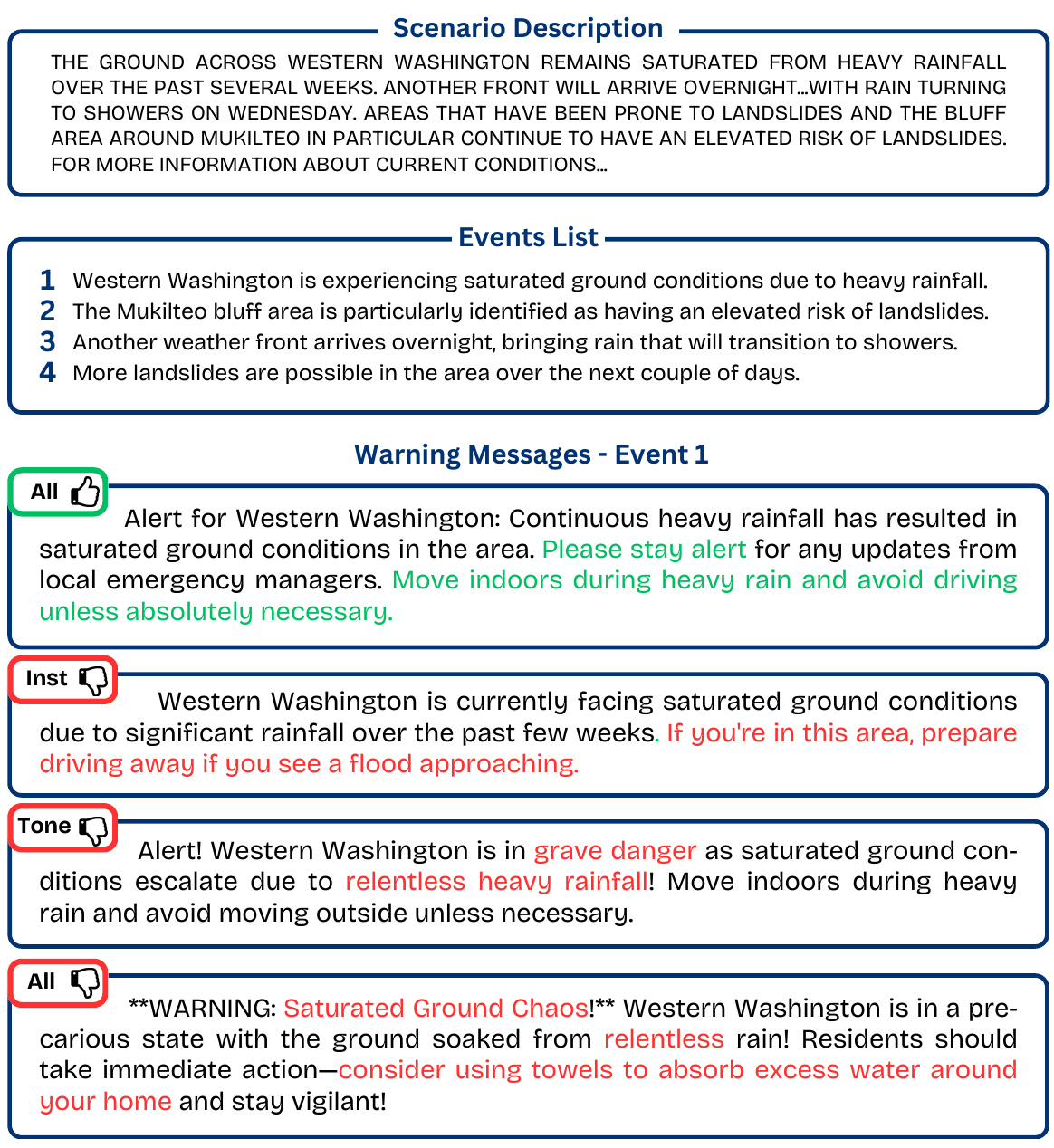}
 \caption{Dataset Entry. For each crisis scenario, a description is provided, then a list of events in chronological order describing that situation. For each event 4 different warning messages are provided (one consistent with expert based guidelines about tone and behavioral instructions and three suboptimal versions).}
 \label{fig:first-page}
\end{figure}

\begin{figure*}[ht!]
 \centering
 \scalebox{0.9}{
 \includegraphics[width=1\linewidth]{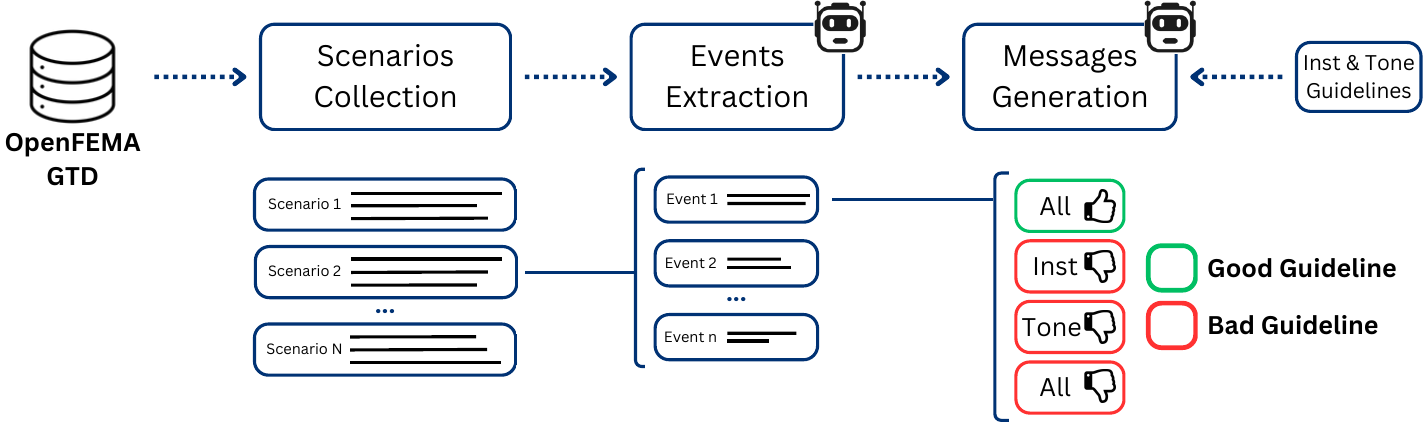}
 }
 \caption{CrisiText Generation Pipeline. Scenarios collection is described in \S \ref{subsec:scenario-collection}, events extraction from scenario description is reported in \S \ref{subsec:scenario-extraction}, and the generation of various warning message versions for each event in \S \ref{subsec:message-generation}.
 }
 \label{fig:generation-pipeline} 
\end{figure*}

To tackle these challenges, Artificial Intelligence (AI) is increasingly making its way as an ally in assisting humans in crisis management \cite{comes2024aicrisisdecisions, harica2024aidisaster, hyun2020aimigrant}. Still, despite the growing number of AI applications, the use of NLP remains limited, mostly focusing on classification tasks \cite{alam2021crisisbench, yin2024crisissense, liu2021crisisbert}. However, with the advent of LLMs it is now possible to address other crucial area of crisis management, i.e. how communication should be structured and delivered \cite{white2011social, hu2016ict}, for example by assisting emergency operators in writing warning messages \cite{otal2024llmassisted}.

As a step in this direction, we introduce \textit{CrisiText}, the first crisis management dataset designed to specialize LLMs on expert-based crisis communication. In Figure \ref{fig:generation-pipeline}, we detail the pipeline employed to develop the dataset. We extracted the scenario descriptions from two existing sources: the FEMA IPAWS Archived Alerts\footnote{\url{https://www.fema.gov/openfema-data-page/ipaws-archived-alerts-v1}} and the Global Terrorism Database (GTD) \cite{lafree2007globalterrorism}. These datasets provide detailed descriptions of various natural disasters and terrorist attacks, respectively. Using GPT-4o-mini,\footnote{specifically, gpt-4o-mini-2024-07-18.} we first derived, for each description, a sequence of textual events in chronological order, simulating the unfolding of the scenario. These events are then utilized to generate warning messages aimed at guiding civilians during and after the specific event. We crafted messages according to two specific dimensions of expert-based guidelines: (i) \textit{Tone} and (ii) \textit{Instructions}. For \textit{Tone}, the messages must ensure proper terminology, provide accurate information, and avoid causing panic. For \textit{Instruction}, instead, FEMA’s guidelines are used to provide grounded suggestions on how to behave depending on the type of crisis. Along with the correctly generated warnings, usable for supervised fine-tuning (SFT), we also produced sub-optimal messages to test preference alignment techniques. These messages may lack lexical correctness, suggestion relevance, or both. Using this approach, we generate over 100,000 warning messages, associated with over 300,000 bad counterparts, covering 13 emergency scenarios. An example from the dataset is shown in Figure \ref{fig:first-page}.

To assess the effectiveness of our dataset for crisis communication, we designed various experimental setups. We conducted experiments with Llama 3 models \cite{dubey2024llama}, using both SFT and ORPO \cite{hong2024orpo}, a preference alignment technique. Specifically, we tuned models to perform the tasks of warning message \textit{generation} and \textit{post-editing}. Results show a comparable performance between ORPO and SFT. Thus, building on the SFT setup, the less computationally demanding method, we run further experiments to test the benefits of providing additional context in warning message generation (i.e. specific instruction guidelines and/or previously generated messages). We analyzed the impact of such additional context both for already seen and out-of-distribution scenario types. The importance of previous messages is evident for already seen scenario types while instructions guidelines are fundamental for out-of-distribution cases. Finally, using the Bad Messages, we fine-tuned an automatic post-editor model which shows promising results in improving the quality of poorly written warning messages.

\section{Related Work}

How to optimally react to crises and emergencies caused by natural or social phenomena has been widely studied \cite{Quarantelli1988, PearsonMitroff1993, Wex2014}. Research in this field focuses on communication practices \cite{HeathOHair2009, FearnBanks2016, SchwarzSeegerAuer2016}, as they are a fundamental component of the crisis management process. In highly dynamic, stressful, and dangerous scenarios, the ability to communicate swiftly and effectively can significantly reduce risks.

In addition to the specific challenges posed by disasters, the rapid evolution of media and communication tools makes information diffusion more complex \cite{HaddowHaddow2022}. In fact, modern technologies such as messaging apps and social media raise opportunities as well as risks. On the one hand, they enable instant delivery of relevant information to a larger audience, thus increasing the likelihood of helping individuals directly involved in such situations. On the other hand, they pose challenges such as information overload \cite{BawdenRobinson2008}, which can create ineffective behaviors and confusion, as well as the proliferation of incorrect information or fake news \cite{Hansson2020,Gisondi2022}.

Within this evolving scenario, AI pipelines are emerging as key tools for managing crises and emergencies, with works exploring domain such as decision-making, mobility studies, and crisis detection \cite{comes2024aicrisisdecisions, hyun2020aimigrant, harica2024aidisaster}.

This field also explores the use of NLP to address practical applications. The diffusion of social networks has allowed the creation of datasets from user-generated content during crisis events like floods or tornadoes \cite{olteanu2014crisislex, imran2016twitter}. Given their tweet-based and human annotations structure, these datasets are mostly used for binary or multi-label classification tasks such as informativeness, crisis detection, or crisis type recognition. These tasks are addressed either by training or fine-tuning deep learning models \cite{alam2021crisisbench, liu2021crisisbert} or using LLMs for zero-shot classification \cite{yin2024crisissense}.

The literature also includes datasets that gathers civilian messages directed to disaster reporting services \cite{munro12dissertation}. This line of research, known as crisis communication, has recently explored the use of LLM-powered systems not only for classifying different types of crises but also for enhancing the efficiency of emergency operators by extracting the most relevant information from a call and display it to the operator to guide the civilian \cite{otal2024llmassisted}.

To fill the gap on warning messages generation within the NLP field, we propose a novel crisis communication dataset covering 13 different scenario types and following expert written guidelines.

\section{\textit{CrisiText} Dataset}
Fine-tuning LLMs for crisis communication requires warning messages that are linked to a sequence of chronologically ordered events. To this end, we identified two existing resources suitable for  creating a crisis communication dataset. OpenFEMA IPAWS Archived Alerts\footnote{\url{https://www.fema.gov/openfema-data-page/ipaws-archived-alerts-v1}} contains warning messages from 2012 to the present issued by over 1,450 alert organizations (mainly for natural disasters). It includes details such as date, time, and geographic information from public emergency alerts across the United States, along with scenario descriptions. The second dataset is the Global Terrorism Database (GTD) \cite{lafree2007globalterrorism}, that focuses on violent attacks providing detailed data on each incident, including date, location, and descriptions.

Since these two resources were not designed for NLP applications nor provided the data or structure we required, we used GPT-4o-mini to generate a synthetic dataset based on them. We first selected those scenarios in which a large population is threatened and then used GPT-4o-mini to derive a list of chronologically ordered events from their descriptions. Then, we prompted the model with guidelines for both effective communication (\textsc{Tone}) and behavioral guidance (\textsc{Instruction}). We thus obtained expert-level warning messages for each event within a scenario.

The following sections will describe in detail each step in the creation of the dataset.

\paragraph{Scenarios Collection.}
\label{subsec:scenario-collection}

Starting from GTD and OpenFEMA's archive, we collected textual descriptions of violent attacks and natural disasters, respectively. To gather these scenarios (textual descriptions) we filtered both datasets according to specific labels. For GTD, we collected scenarios where the focus was on group security rather than incidents involving individuals. Then we filtered scenarios whose description was more than 300 characters, resulting in the collection of \textit{violent attack}, \textit{explosion}, and \textit{arson} scenarios.
For OpenFEMA we focused on collecting scenarios labeled as urgent and linked to FEMA Instruction pages (e.g. \textit{wildfires}, \textit{floods}, \textit{hurricanes}, \textit{thunderstorms}, \textit{landslides}, \textit{tsunamis}, \textit{earthquake}). We used up to 2,000 scenarios for each natural disaster type and included all scenarios from GTD. For more details on this step, see Appendix \ref{app:filtering}.

\paragraph{Events Extraction.}
\label{subsec:scenario-extraction}

To obtain a chronological list of events occurring in each scenario description, we used GPT-4o-mini. In some cases, the descriptions were excessively long (more common in OpenFEMA) or contained unnecessary details (more frequent in GTD). To address these issues in the generation process, OpenFEMA events lists were generated limiting the number of events to maximum 15, while GTD texts underwent an additional generation step: first descriptions were rewritten to eliminate unnecessary details, then the list of maximum 15 events was created. Below, we present the simplified versions of the prompts used to generate the events lists for GTD and OpenFEMA scenarios.\footnote{Quality control and complete prompts are presented in Appendices \ref{app:event-extraction-quality-control} and \ref{app:complete-generation-prompts}, respectively.}

\prompt{STEP 1: Provide a brief overview with the location first, removing unnecessary details like costs, dates, injuries, responsibility, motives, related events, aftermath, and technical info.\\
\\
STEP 2: Create a list of events in the present tense, starting with the location, without identifying individuals. Avoid mentioning vandalism, motivations, suppositions, or related events.}
\prompt{Provide live updates from a natural disaster, focusing only on disaster-related details. List events in real time, starting with impacted locations. Include future event times, and indicate danger when unspecified. Do not add links or current time details.}

\paragraph{Guidelines Collection.}

To accurately control the process of generating warning messages, we collected guidelines addressing message \textsc{Tone} and behavioral \textsc{Instructions}. 

\textsc{Tone} guidelines were gathered from a recent study by \cite{Sutton2019}. In the project, 
the authors conducted a systematic review of the literature on the best practices to deliver effective communication during and in the aftermath of a crisis via short-messaging channels, such as Wireless Emergency Alerts (WEA) or Twitter. Additionally, the authors leveraged recommendations and suggestions from a panel of experts to further refine the results obtained in the first phase of the project (i.e., the systematic review). The panel consisted of 17 individuals with wide expertise in the management of crises and emergencies via WEA and social media, who had experience in situations including floods, earthquakes, hurricanes, terrorism attacks, and active shooter scenarios. The final results of the study identified five main goals that should be pursued in order to deliver effective messages: (i) increasing attention, by designing messages to gather the receiver’s focus; (ii) increasing comprehension, by ensuring clarity, ease of understanding, and explicit mention of threat and location; (iii) ensuring believability, by using a known, trusted source; (iv) enhancing clarity, by avoiding possible ambiguity; and (v) triggering protective action, by clearly and specifically stating the behaviors to adopt in order to reduce the risk of information seeking. Full details about Tone guidelines are available in Section \ref{app:comms_guidelines} of the Appendix.

\textsc{Instructions} Guidelines are extracted from the official FEMA website,\footnote{\url{https://www.ready.gov/}} that provides specific Instructions for each type of crisis event. We collected all the relevant guidelines for the scenarios we gathered, discarding those guidelines not relevant for our dataset (e.g. instructions for specific subcategories such as explosions and power outages).
Then we manually removed the sections that focus on how to prepare before the event, as our task was to assist civilians during and after. 

\paragraph{Warning Messages.}
\label{subsec:message-generation}

After collecting the scenario descriptions and the corresponding events list, we used GPT-4o-mini to generate the warning messages. The model was provided with the full list of events composing each scenario and tasked with producing all the messages at once, so to enhance contextual coherence among messages. Both \textsc{Tone} and \textsc{Instructions} guidelines were included in the prompt to ensure compliance. A simplified prompt for their generation can be found below:\footnote{The complete prompts can be found in Appendix \ref{app:complete-generation-prompts}.}

\prompt{Create warning messages based on the list provided.\\
- Structure: max 300 characters, with clear location and threat, written as standalone updates.\\
- Tone: clear location and threat, keep anonymity, avoid alarming terms, and use clear, simple language. Avoid speculation, links, or unnecessary details.\\
- Instructions: Provide actionable advice from given guidelines \{event\_instructions\}}

Along with the messages generated using the above configuration, referred to as \textit{Good Messages}, we also created their suboptimal counterparts, wich we refers to as \textit{Bad Messages}.
These suboptimal versions were created to (i) explore preference alignment techniques, where a ``chosen'' output is compared to a ``rejected'' one to guide the model toward more suitable generations, and (ii) test post-editing approaches.
The \textit{Bad Messages} were designed by deliberately ignoring or worsening essential aspects of a \textit{good} warning message. Specifically, we generated three variants: messages with poor Tone but correct Instructions (\thumb{Tone}{down}), messages with poor Instructions but correct Tone (\thumb{Inst}{down}), and messages where both aspects were flawed (\thumb{All}{down}).

In total, we generated 100,000 Good Messages, along with an additional 300,000 Bad Messages across nearly 18,000 scenarios spanning 13 types. Each scenario contains roughly 6 events and the average length of a warning message is 255 characters. Table \ref{tab:dataset-information} presents the full list of scenarios and events for each typology.

\begin{table}[htbp]
\centering
\small
\begin{tabular}{lrrrr}
\toprule
\textbf{Type} & \textbf{Scenarios} & \textbf{Events} & \textbf{Events$_\mu$} \\
\midrule
avalanche & 919 & 6,216 & 6.76 \\
attack & 4,412 & 17,003 & 3.85 \\
earthquake & 100 & 347 & 3.47 \\
flood & 1,981 & 13,506 & 6.82 \\
heat & 1,999 & 10,200 & 5.10 \\
hurricane & 1,824 & 13,192 & 7.23 \\
landslide & 146 & 839 & 5.75 \\
thunderstorm & 2,000 & 14,363 & 7.18 \\
tornado & 1,959 & 13,716 & 7.00 \\
tsunami & 130 & 856 & 6.58 \\
volcano & 70 & 477 & 6.81 \\
wildfire & 231 & 1,022 & 4.42 \\
winter weather & 2,000 & 12,717 & 6.36 \\
\midrule
\textbf{Total} & \textbf{17,771} & \textbf{104,454} & \textbf{5.88} \\
\bottomrule
\end{tabular}
\caption{\textit{CrisiText} Datasets statistics.}
\label{tab:dataset-information}
\end{table}

\section{Data Quality}
\label{sec:data_quality}

Once Messages were generated we run two evaluation experiments to assess their quality and suitableness for our purposes. To this end we focused on the Good messages quality, and the perceived difference between Good and Bad messages.\footnote{Further details on setup, prompts, and statistical significance of both experiments are provided in Appendix \ref{app:data-quality}.}

\paragraph{Good Messages Quality.} 
Two crisis communication experts reviewed a subsample of 1,000 examples balanced across the different disaster typologies. As an intrinsic evaluation, we tasked the reviewer with signaling problematic messages and post-editing them. During revision, the experts identified 5 messages (0.5\% of the total) as problematic. These messages included wrong suggestions. Additionally, 38 messages (3.8\% of the total) required minor post-editing (e.g. ``loud explosion'' modified to ``explosion'') with an average HTER \cite{snover2006hter} of 0.11, which is considered negligible \cite{turchi2013coping}. In conclusion, the low number of experts' notes and the low HTER scores can be considered indicators of the good quality of our warning messages.

\paragraph{Good vs Bad messages.}

To assess if the Good and the Bad Messages of \textit{CrisiText} dataset followed our generations guidelines, we employed a double validation process.
For the first step, we set up a pairwise comparison using crowdsourcing, and asked the annotators to select which one among two messages was preferred based on each of the two guideline dimensions (Tone and Instructions). Pairs were created by selecting one good message and one randomly from either \thumb{Tone}{down}, \thumb{Instruct}{down}, or \thumb{All}{down}. After the human evaluation, we repeated the procedure using \texttt{Llama-3.3-70B-Instruct} as a judge. The model was provided with the same evaluation rules and setup, along with the message and events up to that point. The results reported in Table \ref{tab:humans-llm-data-quality} indicate a clear-cut preference for the Good Messages over the Bad Messages with a high consistency between human and LLM judgment (with a Cohen's Kappa of 0.65 and simple agreement of 83.33\%). It should also be noted that when \thumb{Inst}{down} or \thumb{Tone}{down} are evaluated on their own dimension their preference score is dramatically low (as expected), while they received higher preference scores when evaluated on the other dimension (e.g. \thumb{Tone}{down} has higher votes on Inst\textsuperscript{LLM} with respect to Tone\textsuperscript{LLM}). This is expected since \thumb{Tone}{down} is supposed to provide a poor tone but legit instructions (and the converse holds for \thumb{Inst}{down}). 

\begin{table}[t]
\centering
\small
\begin{tabular}{lrrrr}
\toprule
\textbf{Type} & \textbf{Tone\textsuperscript{H}} & \textbf{Tone\textsuperscript{LLM}} & \textbf{Inst\textsuperscript{H}} & \textbf{Inst\textsuperscript{LLM}} \\
\midrule
\thumb{All}{up} & \textbf{98.25\%} & \textbf{100.00\%} & \textbf{98.21\%} & \textbf{100.00\%} \\
\thumb{All}{down} & 1.75\% & 0.00\% & 1.79\% & 0.00\% \\
\midrule
\thumb{All}{up} & \textbf{86.84\%} & \textbf{89.00\%} & \textbf{93.24\%} & \textbf{95.67\%} \\
\thumb{Inst}{down} & 13.16\% & 11.00\% & 6.76\% & 4.33\% \\
\midrule
\thumb{All}{up} & \textbf{95.51\%} & \textbf{99.00\%} &\textbf{80.23\%} & \textbf{73.67\%} \\
\thumb{Tone}{down} & 3.49\% & 1.00\% & 19.77\% & 26.33\% \\
\bottomrule
\end{tabular}
\caption{Comparison of human annotators and LLM-as-a-judge percentage choices. \textsuperscript{H} refers to human annotators, and \textsuperscript{LLM} refers to LLM-as-a-judge. All human choices are statistically significant, with a $p \leq 6.75^{-9}$.}
\label{tab:humans-llm-data-quality}
\end{table}

\section{Experiments}
\label{sec:experiments}

In order to assess the effectiveness of our dataset for crisis communication scenarios, we designed various experimental setups. We simulated real-world tasks to prove that our data can improve models by making them robust under different conditions. Specifically, we fine-tuned models to perform tasks such as warning message generation (also in out-of-distribution scenario types), and warning message post-editing. Additionally, we performed ablation tests to assess the effect of providing the Instruction guidelines and the history of previous messages in these configuration. For all experiments, we used \texttt{Llama-3.1-8B}, testing both the Instruct and Base variants. Details of all the training and generation setups are provided in the Appendix \ref{app:training-arguments}.

\paragraph{Warning message generation.}
This first set of experiments is meant to evaluate the effectiveness of the dataset in supporting warning message generation. To this end, we explored various methodologies: standard Supervised Fine-Tuning (SFT) and a preference alignment paradigm, along with zero-shot and few-shot approaches used as baselines.
The basic prompt, present in all of the configurations, was a short description of the task along with the chain of events of the scenario.

\prompt{Create a warning message informing on the current happening, providing a suggestion, for the last line in the following chain of events (be short, max 300 characters). No other output other than the message.\\
Chain of events: \{chain\_of\_events\}}

Among the various preference alignment options, we selected ORPO \cite{hong2024orpo}, a reference-model-free preference optimization technique. Unlike other alignment techniques, such as DPO \cite{rafailov2024dpo}, ORPO removes the need for an additional alignment phase by integrating it into the fine-tuning phase. While the SFT uses cross-entropy loss, $\mathcal{L}_{SFT}$, ORPO adds the relative ratio loss to the standard one:
\[
\mathcal{L}_{ORPO} = \mathbb{E}_{(x, y_w, y_l)} \left[ \mathcal{L}_{SFT} + \lambda \cdot \mathcal{L}_{OR} \right]
\]
Like all preference alignment algorithms, ORPO requires both a chosen and a rejected output during training. By setting the three variants of Bad Message as rejected, we were able to extend the SFT training to three ORPO setups: ORPO\textsubscript{\thumb{Tone}{down}}, ORPO\textsubscript{\thumb{Inst}{down}}, and ORPO\textsubscript{\thumb{All}{down}}.

\paragraph{Additional Configurations.}
We defined three further configurations adding specific information to the basic prompt to test how the output quality is affected: (i) previous messages from the same scenario, (ii) FEMA Instructions, and (iii) combining both. See appendix \ref{app:training-complete-prompts} for the complete prompts. 

\paragraph{Leave One Scenario Out (LOSO).}
These experiments were conducted to further explore the importance of Instructions guidelines for generalization capabilities of fine-tuned models. We investigated the model's behavior on scenarios that were left out from the training data. To do so, we selected three event types with distinct Instructions (specifically attack, tornado, and winter weather), fine-tuned a model on two of them, and tested on the excluded scenario. To ensure a fair distribution, we focused on scenarios with a single Instruction label and down-sampled the number of elements for each scenario to 1,500, in order to control for the effect of training dataset size.

\paragraph{Post editor.}
The final set of experiments focuses on fine-tuning an LLM specifically for a post-editing task, to correct poorly crafted messages. We also compare its performance with the LLM used in zero-shot post-editing as a baseline. A Bad Message was provided as input and the corresponding Good Message as expected output. In creating such pairs we used a mixture of the three categories of Bad Messages (\thumb{Tone}{down}, \thumb{Inst}{down}, and \thumb{All}{down}), selected through a uniform distribution, ensuring that the dataset contained one-third of each for every Good Message.

\section{Evaluation and Results}

To evaluate the performance of our experiments, we used a combination of traditional metrics and LLM-as-a-judge. The details are provided below.

\paragraph{Metrics.}

We employed overlap metrics to evaluate the similarity between generations produced by our fine-tuned models and the Good Messages. We chose ROUGE1 and ROUGE2 (R1 and R2) \cite{lin2004rouge}, and BLEU (B) \cite{papineni2002bleu}. Although the metrics were developed for machine translation tasks, these metrics help us understand how closely the generation follows the desired structure and terminology. To understand how semantically close are the generations and the gold, we also used BERTScore (BS) \cite{zhang2020bertscore}. BS compares the context embedding of words, providing scores that better align with humans in gen tasks. Along with the traditional metrics, we also employed the LLM-as-a-judge technique introduced in \S \ref{sec:data_quality} to approximate human evaluation. 

\paragraph{Base vs Instruct Models.}

Preliminary results show that there is no significant difference between fine-tuning the Base or the Instruct versions of \texttt{Llama-3.1-8B}. Since their performance is comparable, we chose to continue our experiments with the Instruct model. Full results can be found in Appendix \ref{app:sft-complete-results}.

\begin{table}[ht!]
\centering
\small
\begin{tabular}{lcccccc}
\toprule
\textbf{Setup} & \textbf{R1} & \textbf{R2} & \textbf{B} & \textbf{BS} \\
\midrule
Baseline & 0.305 &  0.104 & 0.083 & 0.675 \\
ORPO & 0.394 & 0.168 & 0.144 & 0.740 \\
SFT & 0.435 & 0.207 & 0.182 & 0.757 \\
SFT\textsubscript{I} & 0.431 & 0.202 & 0.175 & 0.755 \\
SFT\textsubscript{M} & 0.451 & 0.221 & 0.211 & 0.773 \\
SFT\textsubscript{I+M} & \textbf{0.453} & \textbf{0.223} & \textbf{0.213} & \textbf{0.774} \\
\bottomrule
\end{tabular}
\caption{Performance metrics for the various setups. The subscript \textsubscript{I} refers to the incorporation of FEMA Instruction in training, while \textsubscript{M} indicates the use of previous messages. ORPO corresponds to \textsc{Inst}\textsuperscript{\tiny{\faThumbsODown}} and Baseline correspond to Few-shot\textsubscript{C+I}.
}
\label{tab:generation-performances}
\end{table}

\paragraph{Generation Results.}

Getting to the generation experiments, we compared a baseline, the SFT models, and ORPO models. As the baseline, we chose the best-performing setup among the zero-shot and few-shot approaches we tested (see Appendix \ref{app:best-baseline}), while the three ORPO setups did not exhibit substantial differences in terms of automatic metrics (see Appendix \ref{app:orpo-complete-results}). Table \ref{tab:generation-performances} highlights the subpar performance of the baseline, which a qualitative analysis (Appendix \ref{app:zero-shot-sft-analysis}) attributes to the difficulty to follow the  guidelines. With respect to ORPO, SFT achieved better results across all metrics, while the LLM-as-a-judge evaluation does not indicate a clear winner between the two approaches (see Table \ref{tab:orpo-llm-as-a-judge}). Based on these findings, and given the high computational cost of ORPO training, we focused on SFT for the subsequent experiments.

\begin{table}[ht!]
\centering
\small
\begin{tabular}{lrr}
\toprule
\textbf{ORPO Setup} & \textbf{Tone} & \textbf{Instructions} \\
\midrule
\thumb{All}{down} & 51.00\% & 54.67\% \\
\thumb{Instruct}{down} & 50.00\% & 51.33\% \\
\thumb{Tone}{down} & 49.33\% & 52.67\% \\ 
\bottomrule
\end{tabular}
\caption{Percentage of times the LLM-as-a-judge chose the listed ORPO setup instead of the SFT.}
\label{tab:orpo-llm-as-a-judge}
\end{table}

\paragraph{Ablation Experiments.}
Focusing on the different SFT setups, Table \ref{tab:generation-performances} shows that including Instructions during training is not beneficial, while including previous messages improves performance. The latter helps maintaining a consistent message style across all events in the scenario, which is a desirable feature. On the other hand, we hypothesize that the inclusion of Instructions makes the prompt repetitive, potentially having a negative effect on the loss computation during training. A lower loss in this context may lead the model to underfit the data.

In terms of automatic metrics, the best setup for the model combines Instructions and previous messages during fine-tuning, although its performance is comparable to the setup that uses only previous messages. Turning to the LLM-as-a-judge evaluation, presented in Table \ref{tab:sft-llm-as-a-judge}, it is consistent with the results of the automatic metrics. SFT\textsubscript{I+M} is the most frequently selected across both categories. These results are consistent with the qualitative analysis in Appendix~\ref{app:zero-shot-sft-analysis}.

\begin{table}[ht!]
\centering
\small
\begin{tabular}{lcc}
\toprule
\textbf{Setup} & \textbf{Tone} & \textbf{Instructions} \\
\midrule
SFT & 27.00\% & 22.00\% \\
SFT\textsubscript{I} & 24.33\% & 24.67\% \\
SFT\textsubscript{M} & 21.33\% & 25.50\% \\
SFT\textsubscript{I+M} & \textbf{27.33\%} & \textbf{27.83\%} \\
\bottomrule
\end{tabular}
\caption{LLM-as-a-judge results for each configuration.}
\label{tab:sft-llm-as-a-judge}
\end{table}

\paragraph{LOSO.}
Applying automatic metrics to LOSO generations yields only small differences between using FEMA Instructions and not. This can be explained by the fact that generated messages are typically composed of two sentences: the alert (which describes the threat) and the suggestion (which provides instructions on how to respond). Without proper Instruction guidelines, the model can still learn to generate the alert part correctly (which is independent from the FEMA instructions) but struggles with producing accurate behavioral suggestions (which are instead dependent on FEMA instructions). This is confirmed by Table \ref{tab:loso-performance}, which shows a noticeable difference in the suggestion part depending on whether Instructions were used.

\begin{table}[ht]
\centering
\small
\begin{tabular}{lcccc}
\toprule
\textbf{Part \& Setup} & \textbf{R1} & \textbf{R2} & \textbf{B} & \textbf{BS} \\
\midrule
Alert\textsubscript{No Inst} & 0.510 & 0.307 & 0.238 & 0.786 \\
Alert\textsubscript{Inst}& 0.508 & 0.309 & 0.236 & 0.779 \\
Alert\textsubscript{$\Delta$} & \textbf{-0.002} & \textbf{0.001} & \textbf{-0.002} & \textbf{-0.007} \\
\midrule
Sugg\textsubscript{No Inst} & 0.249 & 0.059 & 0.039 & 0.663 \\
Sugg\textsubscript{Inst}& 0.269 & 0.072 & 0.048 & 0.678 \\
Sugg\textsubscript{$\Delta$} & \textbf{0.019} & \textbf{0.013} & \textbf{0.009} & \textbf{0.015} \\ 
\midrule
Total\textsubscript{No Inst} & 0.379 & 0.153 & 0.124 & 0.717 \\
Total\textsubscript{Inst} & 0.390 & 0.163 & 0.132 & 0.726 \\
Total\textsubscript{$\Delta$} & \textbf{0.011} & \textbf{0.010} & \textbf{0.007} & \textbf{0.008} \\
\bottomrule
\end{tabular}
\caption{Performance metrics on different warning message parts of LOSO generations. $\Delta$ represents the difference in metrics between the Inst and No Inst setups.}
\label{tab:loso-performance}
\end{table}

Turning to the LLM-as-a-judge evaluation and differentiating among the three scenario typologies, Table \ref{tab:loso-llm-as-a-judge} clearly shows that the judge perceives a difference between the two training setups. For the excluded scenario, messages generated by the model trained with Instructions are significantly better than those produced by a model trained without them. Notably the Tornado scenario exhibits the smallest difference. This could be explained by the model’s implicit knowledge of specific Instructions before fine-tuning. To confirm this, we used the Perplexity (PPL) metric \cite{arora2016perplexity}. PPL measures how well an LLM predicts the next token in a sequence, indicating how familiar it is with the text. As shown in Table \ref{tab:loso-perplexity}, the Tornado Instructions have the lowest PPL scores, indicating that generating correct suggestions for this scenario was easier even without fine-tuning.

\begin{table}[ht!]
\centering
\small
\begin{adjustbox}{max width=\textwidth}
\begin{tabular}{lrr}
\toprule
\textbf{Excluded Type} & \textbf{No Instructions} & \textbf{Instructions} \\
\midrule
Attack & 26.00\% & \textbf{74.00\%} \\
Tornado & 37.18\% & \textbf{62.82\%} \\ 
Winter Weather & 29.00\% & \textbf{71.00\%} \\
\bottomrule
\end{tabular}
\end{adjustbox}
\caption{ LLM-as-a-judge results on LOSO generations.}
\label{tab:loso-llm-as-a-judge}
\end{table}

\begin{table}[ht!]
\centering
\small
\begin{tabular}{lrr}
\toprule
\textbf{Instruction} & \textbf{PPL (non FT)} & \textbf{PPL (FT)} \\
\midrule
Attack & 8.279 & 9.686 \\
Tornado & 6.697 & 8.177 \\
Winter Weather & 9.086 & 11.050 \\
\bottomrule
\end{tabular}
\caption{Perplexity of \texttt{Llama-3.1-8B}-Instruct before and after fine-tuning on the LOSO setup.
}
\label{tab:loso-perplexity}
\end{table}

\paragraph{Post-Editor.}

Table \ref{tab:post-editor-performance} reports the metrics for the post-editor experiments. We applied automatic metrics to compare all variants of Bad Messages (\thumb{Tone}{down}, \thumb{Instruct}{down}, or \thumb{All}{down}) with their post-edited counterparts, from both the zero-shot and fine-tuned versions of the model. The rationale behind these comparisons is to evaluate if, after the post-editing, the Bad Messages get closer to the Good Messages. The table clearly shows that the fine-tuned version of the model achieves better results than the zero-shot one. The Bad Messages have lower scores on all the metrics, as expected. However, the relatively higher scores of the \thumb{Inst}{down} can be attributed to their generation method, which produces plausible but incorrect suggestions, while maintaining proper communication. The model achieves improvements across all types of Bad Messages, with a cumulative effect on improvement when comparing post-edit the individual fields (Tone and Instructions) vs All. Finally, as shown Table \ref{tab:post-editor-llm-as-a-judge}, The LLM-as-a-judge evaluation confirmed the compelling preference for the post-edited messages.

\begin{table}[ht!]
\centering
\small
\begin{tabular}{llcccccccc}
\toprule
\textbf{Data} & \textbf{R1} & \textbf{R2}& \textbf{B} & \textbf{BS} \\
\midrule
\thumb{All}{down}\textsubscript{or} & 0.266 & 0.092 & 0.066 & 0.619 \\
\thumb{All}{down}\textsubscript{pe-zs} & 0.270 & 0.078 & 0.047 & 0.593 \\
\thumb{All}{down}\textsubscript{pe-ft} & \textbf{0.380} & \textbf{0.154} & \textbf{0.128} & \textbf{0.736} \\
\midrule
\thumb{Tone}{down}\textsubscript{or} & 0.346 & 0.131 & 0.087 & 0.675 \\
\thumb{Tone}{down}\textsubscript{pe-zs} & 0.301 & 0.096 & 0.056 & 0.060 \\
\thumb{Tone}{down}\textsubscript{pe-ft} & \textbf{0.388} & \textbf{0.157} & \textbf{0.132} & \textbf{0.740} \\
\midrule
\thumb{Inst}{down}\textsubscript{or} & 0.359 & 0.168 & 0.141 & 0.695 \\
\thumb{Inst}{down}\textsubscript{pe-zs} & 0.292 & 0.099 & 0.066 & 0.609 \\
\thumb{Inst}{down}\textsubscript{pe-ft} & \textbf{0.405} & \textbf{0.177} & \textbf{0.152} & \textbf{0.747} \\
\bottomrule
\end{tabular}
\caption{Performance metrics of post-editor model. \textit{or} indicates the original message, \textit{pe} the post-edited version, \textit{zs} zero-shot configuration, and \textit{ft} fine-tuned.}
\label{tab:post-editor-performance}
\end{table}

\begin{table}[ht!]
\centering
\small
\begin{adjustbox}{max width=\textwidth}
\begin{tabular}{lrr}
\toprule
\textbf{Message Type} & \textbf{Tone} & \textbf{Instructions} \\
\midrule
Post Edited & \textbf{94.00\%} & \textbf{84.67\%} \\
\thumb{Tone}{down} & 1.00\% & 13.00\% \\
\thumb{Instruct}{down} & 5.00\% & 2.33\% \\
\thumb{All}{down} & 0.00\% & 0.00\% \\
\bottomrule
\end{tabular}
\end{adjustbox}
\caption{LLM-as-a-judge results for post-editing.}
\label{tab:post-editor-llm-as-a-judge}
\end{table}

\section{Conclusions}

In this paper, we presented the first dataset for crisis communication. The dataset covers 13 emergency categories and includes 18,000 crisis scenarios, providing 100,000 events. Each event is associated with one Good Message and three types of Bad Messages, totaling over 400,000 messages. We experimented with Llama 3 models, applying both Supervised Fine-Tuning on Good Messages and a preference alignment technique, ORPO, that also uses Bad Messages. The models were tested on two tasks: warning message \textit{generation} and \textit{post-editing}. Results show that SFT yields better scores than ORPO on overlapping-based metrics, while showed similar performance when evaluated using the LLM-as-a-judge. Further experiments showed how warning message generation improves by providing additional context, i.e. specific guidelines and/or history of previous messages. The importance of adapting to local/different emergency protocols is addressed in LOSO experiments, showing that explicitly including guidelines during training helps the model in adapting to new protocols at inference time. Finally, we fine-tuned an automatic post-editor using Bad Messages, achieving a noticeable improvement in correcting inaccurate messages. We believe that this dataset is a valuable resource for advancing AI-driven crisis communication.

\section*{Limitations}

We emphasize that products based on this dataset should be used as tools to assist humans rather than as a complete replacement for experts, especially when communicating with civilians or dealing with real-world situations. While the dataset has been constructed using SOTA LLMs and expert guidelines, it is important to note that it is synthetic. Some biases from the LLM that generated it are likely present.
Also, even though the generation process follows carefully designed instructions, LLMs are inherently prone to hallucinations. Given the large size of the dataset, the presence of some suboptimal elements is plausible.
Beyond these aspects, we acknowledge that our work focused primarily on the quality of message generation, with limited analysis of factual accuracy, or the potential harmful impact of misleading warning messages. For these reasons, we stress that our approach is meant to be used to assist human experts, not to replace them, as in every sensitive scenario where AI is used.
Additionally, message personalization was not considered in this work, as our focus was on broadcast communication rather than narrowcasting.

\bibliography{custom}

\newpage

\appendix

\section{Scenarios Filtering}
\label{app:filtering}

To obtain a subset of the GTD scenarios characterized by urban situations, such as a terrorist attack in a city center, we stored the scenario descriptions from five different regions: \textit{Australasia \& Oceania}, \textit{East Asia}, \textit{Eastern Europe}, \textit{North America}, and \textit{Western Europe}. Additionally, we focused on five attack types: \textit{Armed Assault}, \textit{Bombing/Explosion}, \textit{Facility/infrastructure Attack}, \textit{Hijacking}, and \textit{Hostage Taking}.

For the OpenFEMA archive, the first step involved collecting entries with the label ``Urgency'' classified as \textit{Expected} or \textit{Immediate}, ``Severity'' categorized as \textit{Extreme}, \textit{Severe}, or \textit{Moderate}, and ``Category'' labeled as \textit{Geo}, \textit{Met}, \textit{Fire}, \textit{Health}, \textit{Env}, or \textit{CBRNE}. In the second step we collected data related to \textit{wildfires}, \textit{floods}, \textit{abnormal heat}, \textit{hurricanes}, \textit{thunderstorms}, \textit{tornadoes} and \textit{winter weather}. Moreover, we extracted \textit{landslides}, \textit{tsunamis} and \textit{volcano} events from the Special Weather category. Additionally, \textit{avalanche} scenarios were separated from \textit{winter weather}. While \textit{earthquake} scenarios were also collected, the majority consisted of \textit{test} messages. Thus, we decided to generate them entirely using synthetic data.

We synthetically generated 100 earthquake events lists by generating scenarios using GPT-3.5-turbo-0125 (Temperature and top P = 1, max tokens = 500, and frequency and presence penalty = 0.5). To add variety to the generated scenarios, we randomized city, magnitude, and the number of events in the scenarios.

\prompt{Generate a dotted list of the live event of an earthquake of magnitude \{magnitude\} for city \{city\}.\\
The list have to start with an earthquake happening.\\
Only gives information about the earthquake.\\
Do not report how the event is felt across the city.\\
DO NOT give indications or suggestions on what to do.\\
The dotted list has to describe the chain of events.\\
Write like you are obtaining information at the moment (like a live report).\\
Do not give narrative details, only focus on events details.\\
Max \{number\_of\_element\} points in the list.}

If the number of events exceeded three, the prompt also asked for information about possible or ongoing aftershocks. Additionally, with a 40\% probability, the prompt included a request to mention evacuation zones.

\section{Full Communication Guidelines}
\label{app:comms_guidelines}
Below, we report the full guidelines as gathered from \cite{Sutton2019}. The study provided information and guidance on how to write effective messages in situations of crisis through a systematic review of the extant literature and the guidance of a panel of experts. We derive five separate goals from the article that should be pursued when aiming to deliver messages that are both useful and effective in reducing negative consequences in the aftermath of situations of emergency.
\paragraph{Increasing attention.}
Short messages should be designed to grab the attention of the message receiver by providing:
\begin{itemize}[noitemsep]
 \item Most up to date and relevant info, along with source
 \item A specific focus on hazard threat and impact, location of incident, instructions about
protective actions to be taken
 \item Sentences using all caps, imperative and directive statements, colors and even
hashtags
 \item Images, when possible
 \item No external links
\end{itemize}

\paragraph{Increasing comprehension.}
Short messages for imminent threats should be easily understandable. Therefore, they must be written to ensure:

\begin{itemize}[noitemsep]
 \item Few (or possibly no) abbreviations, acronyms
 \item No technical jargon
 \item Clear identification of the threat and interested locations
 \item No need for subjective interpretation
 \item Use of maps if possible
\end{itemize}

\paragraph{Ensuring believability.}
A known and trusted source is fundamental to ensure that people will take the message with
the needed attention. Also:
\begin{itemize}[noitemsep]
 \item Alert messages should be used sparingly
 \item They should accurately reflect the seriousness and time of the event
 \item Source, which should be stated early in the message, must be recognizable
 \item Too many alerts decrease believability, the same applies to the perception of
excessive delays between the event and the message
\end{itemize}

\paragraph{Enhancing clarity.}
Short messages should contain unambiguous action phrases that portray the seriousness of
the event and the associated need to take protective measures. So:
\begin{itemize}[noitemsep]
 \item The use of unclear terminology or non-specific location risk information negatively
impacts the public perception of risk
\item The use of words describing the seriousness of the events increase people’s sense
of urgency
\end{itemize}

\paragraph{Triggering Protective Action.}
Short messages for imminent threat should include clear and specific protective action
statements to reduce the urge of people to seek information elsewhere. Hence, the
fundamental elements to be included are:
\begin{itemize}[noitemsep]
    \item Source
    \item Hazard
    \item Location
    \item Guidance
    \item Time
\end{itemize}

\section{Experiments and Data Generation Arguments}
\label{app:training-arguments}

\paragraph{GPT Generation.} 
For dataset generation, we used the OpenAI library to generate text via their API, with the parameters set to a temperature of 1, a top-p of 1, and a maximum token limit of 4096. 

\paragraph{Dataset Splits.}
To train models with our data, we used an 80-10-10 split for the training, development, and test sets. We first collected scenarios based on the main disaster type, then shuffled and split them accordingly.

\paragraph{LoRA Adapter Configuration.}
For all training, we used a LoRA adapter from the \texttt{peft} library on top of the model with the following parameters: a LoRA rank of 16, a LoRA alpha of 16, and a LoRA dropout of 0. The adapters were applied to the following target modules: \texttt{[``q\_proj'', ``k\_proj'', ``v\_proj'', ``o\_proj'', ``gate\_proj'', ``up\_proj'', ``down\_proj'']}. 

\paragraph{SFT.}
We used Hugging Face's \texttt{trl} library with \texttt{SFTTrainer}, configuring the training as follows: a maximum sequence length of 2024, early stopping patience of 5, a learning rate of 3e-4, and a cosine scheduler. The training batch size was 2, the evaluation batch size was 4, and gradients were accumulated over 4 steps. Training ran for 10 epochs with a weight decay of 0.01 and a warmup ratio of 0.03. for the base SFT setup (no previous messages and no FEMA Instructions) we used a lower maximum sequence length of 512 and a higher training batch size of 4. The final model was selected based on the lowest validation loss.

\paragraph{ORPO Training.}
For ORPO training, we used the \texttt{ORPOTrainer} from the \texttt{trl} library. The parameters were the same as those in SFT, except for a maximum sequence length of 512.

\paragraph{LOSO Setup.}
The LOSO setup used a maximum sequence length of 1024, as it did not need to handle excessively long prompts. Unlike other setups, it contained no more than one FEMA Instruction and did not include previous messages.

\paragraph{Post-Editor Setup.}
For the post-editor, we set the maximum sequence length to 600 and the batch size to 4. This setup included only the two messages, without additional context.

\paragraph{Truncation Strategy and Chat Formatting.}
The truncation side was set to left to retain the most relevant content, prioritizing the message to be generated over the instruction or previous messages. The Llama chat template was used to define turns between the assistant and the user.

\paragraph{Llama Generation.}
The generation setup used a temperature and top-p of 1, generating 128 new tokens, as warning messages do not require many tokens.

\paragraph{Evaluation Packages.}
For automatic metrics, we used Hugging Face's \texttt{evaluate} library, importing ROUGE, BLEU, and BERTScore. BERTScore was computed using the \texttt{microsoft/deberta-xlarge-mnli} model to measure similarity.

\section{Event Extraction Quality Control}
\label{app:event-extraction-quality-control}

During the event extraction phase, we tested multiple prompts to minimize the presence of elements that are not standard in broadcast communication. Some of these elements are still present in the dataset. This is not necessarily detrimental, as it helps the model to be more robust in dealing with any information a human operator may consider important. To investigate the quality of the extracted events, we manually reviewed a small subsample of those generated using the final prompt, classifying them into four categories: (i) Good (needs to be communicated), (ii) Neutral (might be communicated if the operator deems it necessary), (iii) Not standard (events connected to the scenario that do not need to be communicated), and (iv) Unrelated (events not connected to the specific scenario). The evaluation returned the following results: 82\% Good, 13\% Neutral, and 5\% Not standard, while there were no Unrelated events. Since the critical aspect was to avoid unrelated events and keep the Not standard ones as low as possible, we considered this sufficient to proceed with the full extraction process.

\begin{table}[ht!]
\centering
\small
\begin{tabular}{cp{5.2cm}}
\toprule
\textbf{Type} & \textbf{Example} \\
\midrule

Good & A severe storm was located 4 miles west of Rutledge, and 6 miles west of Luverne. \\
\midrule
Neutral & Minor hail damage to vehicles expected. \\
\midrule
Not standard & The home of a Catholic woman sustains damages during the storm. \\
\midrule
Unrelated & A wildfire broke out near Santa Rosa, California. \\

\bottomrule
\end{tabular}
\caption{Examples of the four event categories used for quality control.}
\label{tab:event-extraction-examples}
\end{table}

\section{CrisiText Dataset Generation Prompts}
\label{app:complete-generation-prompts}

In this section, we provide the complete prompts for each generation. These were considered the best prompts to guide the generation toward our goals, for both events and messages. 

\paragraph{GTD Scenarios.}

For the first step of event extraction for GTD, we removed unnecessary details, focusing solely on the attack and excluding damages, responsibility claims, and post-attack outcomes, as they are irrelevant to crisis communication.

\prompt{Rewrite the text following these guidelines:\\
- It is important to specify the precise location early in the text.\\
- Remove any reference to cost of damage.\\
- Remove dates and hours indications.\\
- Do not report who and how much are injured.\\
- Do not report claims of responsibility or motivations.\\
- Do not report details happening after the end of the attack.\\
- Do not report linked cases or events prior to the attack.\\
- Do not report technical details.\\
\\
Text:\\
\{scenario["summary"]\}}

In the second step, we extracted lists from the cleaned descriptions, applying additional filtering. We observed that GPT-4o-mini sometimes inferred steps when lacking sufficient information. Since our task does not require strictly factual reporting, this enrichment added useful details and context.

\prompt{Create a dotted list if the following scenario reporting the chain of events of the text.\\
- Use only information of the scenario.\\
- Write to the present.\\
- Each point is a description of its associated event.\\
- Keep anonymity of involved persons but it is important to be precise with the location's name.\\
- Do not report vandalism or motivations for the attack.\\
- Do not report suppositions and related attacks.\\
\\
The event associated to the location have to be specified early.\\
\\
Scenario:\\
\{corrected\_scenario\}}

\paragraph{OpenFEMA Scenarios.}

As with GTD scenarios, we cleaned the OpenFEMA descriptions to remove details irrelevant to public information. We explicitly instructed to specify the threat type for better clarity.

\prompt{Given a description of a natural disaster (\{disaster\}), extract the live updates from it.\\
Make a dotted list (in english) and write as the information are obtained in real time.\\
If not specified, specify what is the danger. Focus on extracting {disaster} information.\\
The dotted list (with - and no indentations) have to describe the chain of event characterizing the event.\\
Time information are reported only when talking about future events.\\
Do not report internet links and present hours.\\
If they are reported, impacted locations have to be at the start of the list.\\
\\
Text:\\
\{event['description']\}}

Some event descriptions were excessively long, leading GPT-4o-mini too extract too long lists. To address this, we forcibly trimmed lists exceeding 15 elements. Below is the prompt used for regenerating the event list.

\prompt{Create a dotted list extracting the most relevant point of the given dotted list (from 5 to 10 points)\\
The list have to describe the chain of events of a natural disaster.\\
\\
Dotted list:\\
\{response\}}

\paragraph{Good Messages.}

Below is the prompt used to generate Good Messages. We refer to dotted lists as numbered lists, as GPT-4o-mini generates messages more reliably when each step is numbered. The prompt includes formatting guidelines, Tone guidelines, and Instruction guidelines for each label of the scenario. If the scenario has $n$ labels:

\prompt{Generate warning messages modifying the numbered list given next, follow these details:\\
- Each message must be numbered (number and a -).\\
- Max 300 character per message.\\
- Create messages for ALL of the specified points.\\
- Location and threat have to be clear in every message.\\
- Keep anonymity of involved persons but it is important to be precise with the location's name in every message.\\
- Modify each item to make it more readable. Warn readers of what is happening live.\\
- Messages do not know the information contained in the next points.\\
- Avoid terminologies that may cause panic in the reader.\\
- Avoid terms such as "terrorist", "dangerous", "extreme", or the type of weapon used.\\
- Each message must offer suggestions on what to do.\\
- For suggestions, find the most relevant information from these documents:\\
For \{guideline\_1\_name\}:\\
\{guideline\_1\_text\}\\
...\\
For \{guideline\_n\_name\}:\\
\{guideline\_n\_text\}\\
\\
Numbered list:\\
\{numbered\_list\}}

\paragraph{Bad Messages.}

For \thumb{Tone}{down}, we used a sensational journalistic style. After several tests, we identified this approach as the most effective way to produce terminology that appears informative while still inducing panic.

\prompt{Generate warning messages modifying the numbered list given next, follow these details:\\
- Each message must be numbered (number and a -).\\
- Max 300 character per message.\\
- Create messages for ALL of the specified points.\\
- Messages do not know the information contained in the next points.\\
{\color{red}- Write in a sensational journalistic style.\\
- Use panic-inducing terminology.\\}
- Each message must offer suggestions on what to do.\\
- For suggestions, find the most relevant information from these documents:\\
For \{guideline\_1\_name\}:\\
\{guideline\_1\_text\}\\
...\\
For \{guideline\_n\_name\}:\\
\{guideline\_n\_text\}\\
\\
Numbered list:\\
\{numbered\_list\}}

For \thumb{Inst}{down}, we aimed to generate suggestions that seemed realistic but, aside from not following FEMA guidelines, contained subtle flaws or impracticalities that would make them ineffective in real situations.

\prompt{Generate warning messages modifying the numbered list given next, follow these details:\\
- Each message must be numbered (number and a -).\\
- Max 300 character per message.\\
- Create messages for ALL of the specified points.\\
- Location and threat have to be clear in every message.\\
- Keep anonymity of involved persons but it is important to be precise with the location's name in every message.\\
- Modify each item to make it more readable. Warn readers of what is happening live.\\
- Messages do not know the information contained in the next points.\\
- Avoid terminologies that may cause panic in the reader.\\
- Avoid terms such as "terrorist", "dangerous", "extreme", and similar.\\
- Each message must offer suggestions.\\
{\color{red}- Suggestions must be realistic and address the situation correctly, but always include flaws, impracticalities, or slightly off advice that still sound plausible to a casual reader.\\
- The advice should be given confidently without any hedging or second-guessing.\\}
\\
Numbered list:\\
\{numbered\_list\}}

For \thumb{All}{down}, we combined both dimensions to further degrade the quality of the messages. To achieve this, we also removed the requirement for suggestions to be somewhat reasonable, resulting in completely ineffective and misleading messages.

\prompt{Generate warning messages modifying the numbered list given next, follow these details:\\
- Each message must be numbered (number and a -).\\
- Max 300 character per message.\\
- Create messages for ALL of the specified points.\\
- Messages do not know the information contained in the next points.\\
{\color{red}- Write in a sensational journalistic style.\\
- Use panic-inducing terminology.\\}
- Each message must offer suggestions.\\
{\color{red}- Suggestions must contain flaws or be slightly off but still sound realistic and plausible to the reader.\\}
\\
Numbered list:\\
\{numbered\_list\}}

\section{Data Quality Assessment}
\label{app:data-quality}

\paragraph{Good Messages Quality Annotation.}
To evaluate Good Messages, the data was organized in a Google Sheet to allow for corrections and provide feedback on each message. In addition to the statistics reported in \S \ref{sec:data_quality}, 19 messages (in the Hurricane and Tornado scenarios) were flagged as containing a specific, correct suggestion added by GPT that was not present in FEMA's guidelines.

\paragraph{Good vs Bad Messages Quality Annotation.} 
Quality was defined in terms of Good messages being perceived as better with respect to Bad Messages. We did not compare Bad Messages against each other since this is not relevant for our work. To perform the data quality check, we selected 5 batches of 6 message pairs, each with the same label typologies: 3 with ``attack'', 1 with ``explosion, attack'', and 1 with ``hurricane''. To conduct this study, we used Prolific,\footnote{\url{https://www.prolific.com/}} a platform that grants fair payment, to select trustworthy annotators. We restricted the pool to native English speakers from the UK and USA with a 100\% acceptance rate. We selected a total of 30 messages, evaluated by 4 annotators each. This subset size was chosen following Hugging Face's LLM-as-a-judge cookbook,\footnote{\url{https://huggingface.co/learn/cookbook/en/llm_judge}} which indicates that 30 samples are sufficient to obtain a good correlation between human judgments and model evaluations.

Forms were created using the Google Forms environment. At the start of each form, FEMA's Instructions for the relevant scenario typologies were provided. Each annotator was presented with the preceding events and the current event characterizing the scenario, along with two warning messages (one from one of the three Bad Messages categories and one Good Message). Annotators were asked to select the message that best aligned with two dimensions: Tone and Instructions. 

The guidelines for the task are detailed below:

\prompt{Thanks for participating in our task. You will be presented with 6 scenarios. Each scenario contains a "current event" that you should focus on and, if available, a list of "previous events". You are asked to evaluate the best message between two options (related to the "current event") based on two criteria:\\
\\
1. **Clarity and Terminology**: The message that is the clearest and uses the most appropriate terminology, while inducing the least amount of panic.\\
\\
2. **Guidelines Adherence**: Based on the provided guidelines, the message that best aligns with them and is most useful in addressing the situation.\\
\\
Please review these criteria carefully before making your choice, as the guidelines may vary across different forms.}

To further confirm the quality, we asked two expert annotators to perform the same evaluation on 50 messages each. This extension allowed us to obtain a total of 220 choices (4 annotators $\times$ 30 instances + 2 annotators $\times$ 50 instances). A binomial test confirmed the significance of the results across both dimensions and for each Bad Message category. Table \ref{tab:good-bad-signifiance} provides a detailed summary of the results.

\begin{table}[ht!]
\centering
\small
\begin{tabular}{lllll}
\toprule
\textbf{Setup} & \textbf{Tone \%} & \textbf{p Tone} & \textbf{Inst \%} & \textbf{p Inst} \\
\midrule
overall & 93.61\% & 5.58e$^{-45}$ & 89.36\% & 6.09e$^{-35}$ \\
\thumb{Inst}{down} & 86.84\% & 1.48e$^{-11}$ & 93.24\% & 9.17e$^{-16}$ \\
\thumb{Tone}{down} & 96.51\% & 1.37e$^{-21}$ & 80.23\% & 6.75e$^{-09}$ \\
\thumb{All}{down} & 98.25\% & 4.02e$^{-16}$ & 98.21\% & 7.91e$^{-16}$ \\
\bottomrule
\end{tabular}
\caption{Percentage of \textsc{All}\textsuperscript{\tiny{\faThumbsOUp}}-preferred responses and corresponding p-values from binomial tests for tone and guideline dimensions.}
\label{tab:good-bad-signifiance}
\end{table}

\paragraph{LLM as a Judge.} 
For the LLM-as-a-judge, we provided the same elements as those given to human annotators: FEMA's Instructions, the previous events, the current event, and the two messages. To minimize confusion, the evaluation of the two dimensions was conducted in two separate inference iterations. The prompts used for the evaluations are provided below.

\prompt{You are an expert evaluator tasked with comparing two warning messages.\\
\\
- **Scenario:** These messages were generated for {type}.\\
Messages are generated for the last one of the following events:\\
{events}\\
- **Purpose:** The messages aim to inform the public.\\
\\
**Messages to Compare:**\\
- **Message A:** {message\_a}\\
- **Message B:** {message\_b}\\
\\
**Your Task:**\\
Compare the messages based on how well they balance clarity and tone. Evaluate which message:\\
1. Is clearer and easier to understand.\\
2. Uses calm, neutral language that avoids panic-inducing terms.\\
Identify the message that achieves the best tradeoff between these two aspects.\\
\\
At the end report your choice, exactly in this format "Choice: [A/B]."}

\prompt{You are an expert evaluator tasked with comparing two warning messages.\\
\\
- **Scenario:** These messages were generated for {type}.\\
Messages are generated for the last one of the following events:\\
{events}\\
- **Purpose:** The messages aim to inform the public.\\
\\
**FEMA Guidelines for this Scenario:**\\
{Instructions}\\
\\
**Messages to Compare:**\\
- **Message A:** {message\_a}\\
- **Message B:** {message\_b}\\
\\
**Your Task:**\\
Compare the messages based on their adherence to FEMA guidelines. Evaluate which message better aligns with the provided guidelines and delivers the most appropriate advice.\\
\\
At the end report your choice, exactly in this format "Choice: [A/B]."}

\section{Training Prompts}
\label{app:training-complete-prompts}

The prompt template that we used in our training setups, with all optional blocks shown in curly brackets, is:

\prompt{Based on the provided guidelines, Create a warning message informing on the current happening, providing a suggestion, for the last line in the following chain of events (be short, max 300 characters). No other output other than the message.\\
\\
\{Guidelines Block\}\\
\{Previous messages Block\}\\
\\
Chain of events: \{chain\_of\_events\}}

The Guidelines block contains the FEMA Instructions relative to every labels characterizing the crisis scenario. If a scenario has $n$ labels the block takes the form:

\prompt{Guidelines:\\
For \{guideline\_1\_name\}:\\
\{guideline\_1\_text\}\\
...\\
For \{guideline\_n\_name\}:\\
\{guideline\_n\_text\}}

 The Previous Messages Block includes the messages corresponding to earlier events in the same scenario. When generating the $n$-th message of a scenario, the block takes the form:

\prompt{Previous messages:\\
\{previous\_message\_1\}\\
...\\
\{previous\_message\_(n-1)\}}

The base prompt, without additional blocks, is used for SFT and all ORPO configurations, while SFT\textsubscript{I}, SFT\textsubscript{M}, and SFT\textsubscript{I+M} use combinations of the optional blocks.

\section{SFT Complete Results}
\label{app:sft-complete-results}

Table \ref{tab:full-standard-performance} reports the performance of the fine-tuned Base and Instruct models.

\begin{table}[ht!]
\centering
\small
\begin{tabular}{llcccccc}
\toprule
\textbf{Model} & \textbf{Setup} & \textbf{R1} & \textbf{R2} & \textbf{B} & \textbf{BS} \\
\midrule
Base & SFT & 0.432 & 0.206 & 0.182 & 0.758 \\
Base & SFT\textsubscript{I} & 0.336 & 0.336 & 0.177 & 0.760 \\
Base & SFT\textsubscript{M} & 0.353 & 0.353 & 0.204 & \textbf{0.775} \\
Base & SFT\textsubscript{I+M} & \textbf{0.450} & \textbf{0.223} & \textbf{0.213} & 0.773 \\
\midrule
Instruct & SFT & 0.435 & 0.207& 0.182 & 0.757 \\
Instruct & SFT\textsubscript{I} & 0.431 & 0.202 & 0.175 & 0.755 \\
Instruct & SFT\textsubscript{M} & 0.451 & 0.221 & 0.211 & 0.773 \\
Instruct & SFT\textsubscript{I+M} & \textbf{0.453} & \textbf{0.223} & \textbf{0.213} & \textbf{0.774} \\
\bottomrule
\end{tabular}
\caption{Performance metrics for various training setups. The subscript \textsubscript{I} refers to the incorporation of Instruction in training, while \textsubscript{M} indicates the use of previous messages in training.}
\label{tab:full-standard-performance}
\end{table}

\section{Best Baseline}
\label{app:best-baseline}

The baseline presented in Table \ref{tab:generation-performances} represents the best-performing one, identified across multiple zero-shot and few-shot configurations. For both of these techniques, we computed automatic metrics over three configurations: (i) without additional information, (ii) with communication guidelines and FEMA Instructions, and (iii) with communication guidelines, FEMA Instructions, and previous messages. The prompt template that we used, with all optional blocks shown in curly brackets, is:

\prompt{Based on the provided guidelines, Create a warning message informing on the current happening, providing a suggestion, for the last line in the following chain of events (be short, max 300 characters). No other output other than the message.\\
\\
\{Guidelines Block\}\\
\{Examples Block\}\\
\{Previous messages Block\}\\
\\
Chain of events: \{chain\_of\_events\}}

To switch from a zero-shot to a few-shot setup, we added the Example Block. This block contains two example chains of events and their corresponding output messages, chosen according to the label of each scenario. The format is:

\prompt{Example:\\
Chain of events:
\{chain\_of\_events\_for\_labels\_eg\_1\}\\
Message: \{message\_for\_labels\_eg\_1\}\\
Chain of events:
\{chain\_of\_events\_for\_labels\_eg\_2\}\\
Message: \{message\_for\_labels\_eg\_2\}\\
End of example}

The Guidelines block consists of two parts: fixed communication guidelines (the same used during the generation of the dataset \ref{app:complete-generation-prompts}) and FEMA Instructions (as we did in the training prompts \ref{app:training-complete-prompts}). The full block is reported below:

\prompt{Communication guidelines:\\
- Location and threat have to be clear in every message.\\
- Keep anonymity of involved persons but it is important to be precise with the location's name in every message.\\
- Modify each item to make it more readable. Warn readers of what is happening live.\\
- Avoid terminologies that may cause panic in the reader.\\
- Avoid terms such as ``terrorist'', ``dangerous'', ``extreme'', and similar.\\
- Each message must offer suggestions on what to do.\\
\\
Guidelines:\\
For \{guideline\_1\_name\}:\\
\{guideline\_1\_text\}\\
...\\
For \{guideline\_n\_name\}:\\
\{guideline\_n\_text\}}

Finally, the Previous Messages Block is identical to that described in Appendix \ref{app:training-complete-prompts}.

\prompt{Previous messages:\\
\{previous\_message\_1\}\\
...\\
\{previous\_message\_(n-1)\}}

In Table \ref{tab:zero-few-shot-gen-performances}, we report the automatic metric scores for each setup. Overall, few-shot configuration achieve better performances compared to the zero-shot ones. We chose Few-shot\textsubscript{C+I} as our reference baseline since it shows the best overall performance, having the higher R1 and R2, and the second best BS.

\begin{table}[ht!]
\centering
\small
\begin{tabular}{lcccc}
\toprule
\textbf{Setup} & \textbf{R1} & \textbf{R2} & \textbf{B} & \textbf{BS} \\
\midrule
Zero-shot & 0.248 & 0.075 & 0.048 & \textbf{0.686} \\
Zero-shot\textsubscript{C+I} & \textbf{0.304} & \textbf{0.098} & 0.060 & 0.618 \\
Zero-shot\textsubscript{C+I+M} & 0.296 & 0.095 & \textbf{0.074} & 0.645 \\
\midrule
Few-shot & 0.299 & \textbf{0.104} & 0.085 & \textbf{0.692} \\
Few-shot\textsubscript{C+I} & \textbf{0.305} & \textbf{0.104} & 0.083 & 0.675 \\
Few-shot\textsubscript{C+I+M} & 0.299 & 0.101 & \textbf{0.088} & 0.672 \\
\bottomrule
\end{tabular}
\caption{Performance metrics for the various setups. Subscript \textsubscript{C} indicates the use of communication guidelines, \textsubscript{I} refers to the incorporation of FEMA Instruction, and \textsubscript{M} denotes the inclusion of previous messages.}
\label{tab:zero-few-shot-gen-performances}
\end{table}

\section{ORPO Complete Results}
\label{app:orpo-complete-results}

Table \ref{tab:full-orpo-performance} presents the complete results of all ORPO training setups.

\begin{table}[htb]
\centering
\small
\begin{tabular}{llcccccccc}
\toprule
\textbf{Model} & \textbf{Setup} & \textbf{R1} & \textbf{R2} & \textbf{B} & \textbf{BS} \\
\midrule
Base & \thumb{All}{down} & 0.368 & 0.148 & 0.124 & 0.717 \\
Base & \thumb{Tone}{down} & 0.361 & 0.139 & 0.116 & 0.718 \\
Base & \thumb{Inst}{down} & \textbf{0.371} & \textbf{0.148} & \textbf{0.122} & \textbf{0.722} \\
\midrule
Instruct & \thumb{All}{down} & 0.389 & 0.166 & 0.142 & 0.738 \\
Instruct & \thumb{Tone}{down} & 0.390 & 0.167 & 0.142 & 0.739 \\
Instruct & \thumb{Inst}{down} & \textbf{0.394} & \textbf{0.168} & \textbf{0.144} & \textbf{0.740} \\
\bottomrule
\end{tabular}
\caption{Performance metrics of ORPO training with various types of Bad Messages.}
\label{tab:full-orpo-performance}
\end{table}

\section{LOSO Complete results}
\label{app:loso-complete-results}

Complete results for the LOSO experiments are reported in Table \ref{tab:full-loso-performance}. Notice that, even on these metrics, the Tornado typology does not show major differences between fine-tuning with Instructions and without Instructions.

\begin{table}[ht!]
\centering
\small
\begin{tabular}{p{0.6cm}p{1.2cm}cccc}
\toprule
\textbf{Type} & \shortstack{\textbf{Part \&}\\\textbf{Setup}} & \textbf{R1} & \textbf{R2} & \textbf{B} & \textbf{BS} \\
\midrule

\multirow{9}{*}{\rotatebox[origin=c]{90}{Attack}}
 & Alert\textsubscript{No Inst} & 0.458 & 0.257 & 0.189 & 0.774 \\
 & Alert\textsubscript{Inst}    & 0.462 & 0.261 & 0.184 & 0.768 \\
 & Alert\textsubscript{$\Delta$} & \textbf{0.003} & \textbf{0.004} & \textbf{-0.005} & \textbf{-0.006} \\
\cmidrule(lr){2-6}
 & Sugg\textsubscript{No Inst}  & 0.261 & 0.066 & 0.046 & 0.673 \\
 & Sugg\textsubscript{Inst}     & 0.296 & 0.089 & 0.060 & 0.682 \\
 & Sugg\textsubscript{$\Delta$} & \textbf{0.035} & \textbf{0.023} & \textbf{0.014} & \textbf{0.010} \\
\cmidrule(lr){2-6}
 & Total\textsubscript{No Inst} & 0.370 & 0.139 & 0.113 & 0.722 \\
 & Total\textsubscript{Inst}   & 0.384 & 0.151 & 0.118 & 0.724 \\
 & Total\textsubscript{$\Delta$} & \textbf{0.014} & \textbf{0.012} & \textbf{0.005} & \textbf{0.002} \\
\midrule

\multirow{9}{*}{\rotatebox[origin=c]{90}{Tornado}}
 & Alert\textsubscript{No Inst} & 0.526 & 0.315 & 0.255 & 0.783 \\
 & Alert\textsubscript{Inst}    & 0.510 & 0.306 & 0.246 & 0.769 \\
 & Alert\textsubscript{$\Delta$} & \textbf{-0.016} & \textbf{-0.009} & \textbf{-0.010} & \textbf{-0.014} \\
\cmidrule(lr){2-6}
 & Sugg\textsubscript{No Inst}  & 0.271 & 0.071 & 0.049 & 0.675 \\
 & Sugg\textsubscript{Inst}     & 0.272 & 0.071 & 0.050 & 0.686 \\
 & Sugg\textsubscript{$\Delta$} & \textbf{0.001} & \textbf{0.000} & \textbf{0.001} & \textbf{0.011} \\
\cmidrule(lr){2-6}
 & Total\textsubscript{No Inst} & 0.389 & 0.157 & 0.134 & 0.721 \\
 & Total\textsubscript{Inst}    & 0.390 & 0.157 & 0.138 & 0.726 \\
 & Total\textsubscript{$\Delta$} & \textbf{0.001} & \textbf{0.000} & \textbf{0.004} & \textbf{0.005} \\
\midrule

\multirow{9}{*}{\rotatebox[origin=c]{90}{Winter Weather}}
 & Alert\textsubscript{No Inst} & 0.545 & 0.350 & 0.271 & 0.801 \\
 & Alert\textsubscript{Inst}    & 0.552 & 0.359 & 0.278 & 0.801 \\
 & Alert ($\Delta$)& \textbf{0.007} & \textbf{0.009} & \textbf{0.007} & \textbf{0.000} \\
\cmidrule(lr){2-6}
 & Sugg\textsubscript{No Inst}  & 0.215 & 0.040 & 0.022 & 0.641 \\
 & Sugg\textsubscript{Inst}     & 0.238 & 0.056 & 0.033 & 0.666 \\
 & Sugg\textsubscript{$\Delta$} & \textbf{0.023} & \textbf{0.016} & \textbf{0.011} & \textbf{0.026} \\
\cmidrule(lr){2-6}
 & Total\textsubscript{No Inst} & 0.378 & 0.163 & 0.126 & 0.709 \\
 & Total\textsubscript{Inst}    & 0.397 & 0.180 & 0.139 & 0.727 \\
 & Total\textsubscript{$\Delta$} & \textbf{0.019} & \textbf{0.017} & \textbf{0.013} & \textbf{0.018} \\
\bottomrule
\end{tabular}
\caption{Full LOSO results divided by event typology. $\Delta$ represents the difference in metrics between the Inst and No Inst setups.}
\label{tab:full-loso-performance}
\end{table}

\section{Generation Qualitative Analysis}
\label{app:zero-shot-sft-analysis}

To gain better insights on the characteristics of the generated warning messages, we performed a qualitative analysis comparing the main approaches we used: zero-shot, few-shot, and SFT. In this analysis we included the best- vs worst-performing setups of each technique. The chosen worst setups are Zero-shot, Few-shot, and SFT, while the best ones are Zero-shot\textsubscript{C+I}, Few-shot\textsubscript{C+I}, and SFT\textsubscript{I+M}.

\begin{table}[ht]
\centering
\small
\begin{tabular}{lcccc}
\toprule
\textbf{Generation} & \textbf{Mean} & \textbf{Min} & \textbf{Max} & \textbf{Median} \\
\midrule
Gold & 263.55 & 179 & 357 & 262.5 \\
\midrule
SFT & 264.59 & 159 & 354 & 264.0 \\
SFT\textsubscript{I+M} & 268.77 & 167 & 335 & 276.0 \\
\midrule
Zero-shot & 182.88 &  71 & 309 & 177.5 \\
Zero-shot\textsubscript{C+I} & 391.19 & 127 & 610 & 386.0 \\
Zero-shot\textsubscript{C+I+M} & 321.96 &  78 & 540 & 316.5 \\
\midrule
Few-shot & 252.76 & 143 & 487 & 249.0 \\
Few-shot\textsubscript{C+I} & 302.97 &  90 & 482 & 304.5 \\
Few-shot\textsubscript{C+I+M} & 318.52 & 144 & 664 & 304.5 \\
\bottomrule
\end{tabular}
\caption{Statistics of character counts across the different generation methods.}
\label{tab:zero-few-shot-gen-lengths}
\end{table}

\paragraph{SFT vs SFT\textsubscript{I+M}.}
There are no significant differences between the two: both succeed in producing non–panic-inducing messages that are informative and provide correct advice contextualized to the situation. The terminology is generally better in the SFT\textsubscript{I+M} setup than in the SFT one, likely due to greater exposure during training to similar messages and FEMA Instructions (since the best setup includes previous messages and FEMA texts).

\paragraph{Zero-shot vs Zero-shot\textsubscript{C+I}.}
Both version have poor performances for different reasons. The Zero-shot setup produces very short messages that contain missing/incomplete event information and/or recommendations. Furthermore messages contain panic inducing terminology. Both problems can be explained by the absence of any guideline. The Zero-shot\textsubscript{C+I}, on the other hand, generates messages with more complete information about both the events and the recommendations. However, it produces excessively long messages often using a journalistic reporting style. Moreover, the recommendations are not always correct, as they sometimes fail to warn about the current event (a problem that also occurs in the Few-shot setups, as discussed in the next paragraph).

\paragraph{Few-shot vs Few-shot\textsubscript{C+I}.}
In these experiments, the generated messages show better adherence to the desired style compared to the zero-shot setups: short, with compliant suggestions, and non–panic-inducing terminology. This improvement is mostly due to the presence of examples in the prompt. There are no major differences between the two setups, as also confirmed by the automatic metrics in Table \ref{tab:zero-few-shot-gen-performances}. Still, given the absence of guidelines, Few-shot setup is sometimes prone to generating messages deviating from the expected message format. The main issue with both setups is their tendency to fail to generate a message for the last element in the chain of events (the one that should be communicated). Instead, they often produce a message for another event or a summary of a previous one, which can lead to the omission of important details that needed to be communicated.

\paragraph{SFT\textsubscript{I+M} vs Zero-shot\textsubscript{C+I} vs Few-shot\textsubscript{C+I}.}
Based on the previous insights and analysis, we conclude that SFT is the most suitable option for the Warning Message Generation task, as it produces reliable and informative outputs, which are essential for this application. The other two techniques cannot be considered reliable in these aspects. \\

Table \ref{tab:base-vs-best-generation-techniques} presents a comparison of the warning messages generated by the 6 techniques discussed above (SFT, SFT\textsubscript{I+M}, Zero-shot, Zero-shot\textsubscript{C+I}, Few-shot and Few-shot\textsubscript{C+I}), while Table \ref{tab:zero-vs-few-vs-sft-comparison} presents a comparison between the three best-performing setups. Finally Table \ref{tab:zero-few-shot-gen-lengths} presents the generation lengths of the various techniques, supporting the discussion in this section.

\section{Data Sources Licensing and Citation}

Original GTD descriptions are not redistributed in this dataset. Only derived or transformed information has been included.

\begin{quote}
\textbf{START (National Consortium for the Study of Terrorism and Responses to Terrorism).} (2022). \textit{Global Terrorism Database, 1970--2020 (May 2022 Release).} University of Maryland. \url{https://www.start.umd.edu/data-tools/GTD}
\end{quote}

FEMA data are public domain and redistributed according to FEMA’s open data policy.

\begin{quote}
\textbf{Federal Emergency Management Agency (FEMA).} (2024). \textit{OpenFEMA Dataset: IPAWS Archived Alerts - v1.} Retrieved from \url{https://www.fema.gov/api/open/v1/IpawsArchivedAlerts} on 21 May 2024, 16:32.
\end{quote}

\textit{This product uses the Federal Emergency Management Agency’s OpenFEMA API, but is not endorsed by FEMA. The Federal Government or FEMA cannot vouch for the data or analyses derived from these data after the data have been retrieved from the Agency's website(s).}

\begin{table*}[ht]
\tiny
\centering
\begin{tabularx}{\textwidth}{>{\raggedright\arraybackslash}X
                              >{\raggedright\arraybackslash}X
                              >{\raggedright\arraybackslash}X
                              >{\raggedright\arraybackslash}X}
\toprule
\multicolumn{1}{c}{\textbf{Chain of Events}} & \multicolumn{1}{c}{\textbf{SFT\textsubscript{I+M}}} & \multicolumn{1}{c}{\textbf{SFT}} & \multicolumn{1}{c}{\textbf{Gold}} \\
\midrule
-Nebraska: Nemaha, Richardson, and Pawnee Counties are under a heat advisory.

-Iowa: Montgomery, Page, and Fremont Counties are experiencing extreme heat conditions.

-Heat index values are expected to reach 109 today and 106 on Tuesday. &
A heat index of 109 is expected in your area for today, and it's set to drop to 106 on Tuesday. It is crucial to stay cool and hydrated. Consider taking cool showers or baths if you feel overheated. Remember to check on elderly neighbors and family members, ensuring they are safe. & 
**Heat Wave Alert for Today and Tuesday:** Expect heat index values soaring to 109 degrees today and up to 106 degrees on Tuesday. To protect yourself, avoid strenuous activities outdoors, drink plenty of fluids, and keep an eye on those who might need extra care during the heat. & 
Today's heat index values are forecasted to reach 109°F, with 106°F expected on Tuesday. It’s important to stay cool and hydrated. Avoid strenuous activities and take breaks indoors where possible to minimize heat-related health concerns. \\
\midrule
-Assailants approach the T-Mobile mobile tower located in Lodz, Poland.

-The assailants initiate a fire at the mobile tower.

-The fire spreads, affecting the structure of the tower. 

-Emergency services are alerted and respond to the scene.

-Firefighters work to extinguish the flames and secure the area. &
**Report: Firefighters at T-Mobile Tower, Lodz, Poland.** Firefighters are currently working to extinguish flames. If you are nearby, please stay clear of the area to allow responders to do their jobs. Monitor official updates for any directions and avoid unnecessary risks. &
Firefighters are actively battling the flames at the T-Mobile mobile tower in Lodz. If you are in the area, it’s vital to stay clear to allow a smooth response. Remember to keep your phone handy for emergency updates and stay indoors if you observe smoke. &
**Report: Firefighters combating flames at T-Mobile Tower, Lodz, Poland.** Fire crews are actively working to extinguish the blaze and secure the surrounding area. Stay away from the site and keep a safe distance. If you observe any dangerous smoke, find safety in a well-ventilated space away from the affected zone. \\
\toprule
\multicolumn{1}{c}{\textbf{Chain of events}} & \multicolumn{1}{c}{\textbf{Zero-shot\textsubscript{C+I}}} & \multicolumn{1}{c}{\textbf{Zero-shot}} & \multicolumn{1}{c}{\textbf{Gold}} \\
\midrule
-MELBOURNE, PALM BAY, COCOA BEACH are currently affected by the hurricane.

-High threat to life and property due to wind remains steady; peak wind forecasted at 40-50 MPH with gusts up to 65-70 MPH.

-Window for tropical storm force winds is until late evening; dangerous winds of 74 to 110 MPH may occur.

-Residents are urged to seek shelter immediately to avoid serious injury or loss of life due to hazardous winds.

-Life-threatening storm surge possibly affecting locations with 2-4 feet inundation forecasted above ground in surge-prone areas. &
**URGENT - MELBOURNE, PALM BAY, COCOA BEACH AREA**: Tropical storm-force winds expected until late evening, {\color{red}with peak winds forecasted at 40-50 MPH and gusts up to 65-70 MPH.} Life-threatening storm surge may inundate coastal areas with 2-4 feet of floodwater. **Seek shelter IMMEDIATELY to avoid serious injury or loss of life.** {\color{red}If safe to do so, turn off main breaker or fuse box to prevent electric shock} and protect yourself and your property. Only use phone for emergencies and stay away from w &
URGENT - SEEK SHELTER IMMEDIATELY if in MELBOURNE, PALM BAY, or COCOA BEACH. Danger of life-threatening storm surge {\color{red}and winds of 74-110 MPH persists}. Stay indoors until the danger has passed. &
Storm surge threats are on the rise, with expected inundation of 2-4 feet in surge-prone areas. Move to high ground or designated shelter locations immediately to minimize risk. \\
\midrule
-Pearl River Near Rockport: Flood Warning continues.

-Pearl River Near Monticello: Flood Warning continues.

-Pearl River Near Columbia: Flood Warning continues with minor flooding occurring. &
**{\color{red}Breaking News:} Pearl River Near Rockport - Ongoing Flood Warning**

Flooding continues along the Pearl River in Rockport, threatening homes and roads. Residents advised to **stay informed and avoid driving except in emergencies**. Stay tuned to local authorities for updates. &
Flooding persists along the Pearl River. Consider postponing outdoor activities and staying updated on weather forecasts for updates. &
**Pearl River Near Columbia**: A flood warning is ongoing with minor flooding reported. Those in the vicinity should remain in safe, dry locations and prepare emergency supplies. Keep an ear out for updates from local authorities and remember: do not attempt to cross flooded roads. \\
\toprule
\multicolumn{1}{c}{\textbf{Chain of events}} & \multicolumn{1}{c}{\textbf{Few-shot\textsubscript{C+I}}} & \multicolumn{1}{c}{\textbf{Few-shot}} & \multicolumn{1}{c}{\textbf{Gold}} \\
\midrule
-In Kuqa, Xinjiang Uygur, China, a group of militants approaches a public security bureau using a tricycle loaded with explosives.

-One militant engages in a confrontation with police and is subsequently shot.

-Another militant detonates explosives at the location.

-Two additional militants are apprehended by authorities at the scene. &
~{\color{red}Attention, residents near the public security bureau in Kuqa, Xinjiang Uygur, China: Be advised that an explosion has occurred at the location after a confrontation between police and suspected individuals.} Remain at a safe distance and follow authorities' instructions for further guidance. &
Be advised, a dangerous situation is unfolding in Kuqa, Xinjiang Uygur, China. The scene is not secure. Please avoid approaching the area, follow instructions from law enforcement, and exercise extreme caution. Stay indoors for now and monitor local news for further updates. &
- **Update for Kuqa, Xinjiang Uygur**: Authorities have arrested two individuals related to the incident at the public security bureau. Stay clear of the area to allow emergency services to operate efficiently. Listen for updates from local officials and follow their instructions to maintain safety. \\
\midrule
-Hurricane Warning issued for Morgan City, Patterson, and Franklin, with hurricane-force winds expected within 36 hours.

-Storm Surge Warning alerts for life-threatening inundation, with potential inundation of 6-9 feet above ground.

-Peak Wind Forecast indicates winds of 50-65 mph with gusts up to 85 mph; potential for increase to over 110 mph.

-Immediate preparations and emergency plans are needed to mitigate extreme wind and storm surge damage. &
**Critical Update for Morgan City, Patterson, and Franklin:** {\color{red}Life-threatening hurricane-force winds and surges are predicted within 36 hours, with winds up to 110 mph and 6-9 feet of flooding expected.} Please evacuate if instructed and go to pre-arranged shelter locations or shelter in place according to local guidance to ensure your safety. &
**Prepare for Hurricane Landfall:** {\color{red}A hurricane-force storm is approaching, with wind speeds reaching 50-65 mph and gusts over 85 mph. Expect significant storm surge inundation of 6-9 feet above ground through Thursday.} Take immediate action to secure property, move to higher ground, and have a plan for emergency shelter and communication. &
Immediate action is essential to reduce risks from wind and storm surge. Gather emergency supplies, including water and food. Consider safe shelter options such as an interior room or storm cellar, and stay tuned to authorities for ongoing updates. \\
\bottomrule
\end{tabularx}
\caption{Comparison of worst and best setup for SFT, Zero-shot, and Few-shot. Text highlighted in red indicates example of the problems discussed in the Appendix \ref{app:zero-shot-sft-analysis}.}
\label{tab:base-vs-best-generation-techniques}
\end{table*}

\FloatBarrier

\begin{table*}[ht]
\small
\centering
\begin{tabularx}{\textwidth}{>{\raggedright\arraybackslash}X 
                              >{\raggedright\arraybackslash}X 
                              >{\raggedright\arraybackslash}X
                              >{\raggedright\arraybackslash}X}
\toprule
\multicolumn{1}{c}{\textbf{Chain of events}} & \multicolumn{1}{c}{\textbf{Zero-shot\textsubscript{C+I}}} & \multicolumn{1}{c}{\textbf{Few-shot\textsubscript{C+I}}} & \multicolumn{1}{c}{\textbf{SFT\textsubscript{I+M}}} \\
\midrule
-Nebraska: Nemaha, Richardson, and Pawnee Counties are under a heat advisory.

-Iowa: Montgomery, Page, and Fremont Counties are experiencing extreme heat conditions.

-Heat index values are expected to reach 109 today and 106 on Tuesday. &
**HEAT ADVISORY ALERT - Omaha, NE \& Council Bluffs, IA**

Residents of Nemaha, Richardson, and Pawnee Counties in Nebraska, as well as Montgomery, Page, and Fremont Counties in Iowa, are advised to take precautions in the extreme heat. Temperatures are expected to reach 109 degrees today and 106 on Tuesday. 

Stay cool and hydrated by drinking plenty of fluids, taking cool showers, and finding shade. If no air conditioning is available, consider visiting a cooling center. Be cautious of heat-related illnesses and check in on vulnerable neighbors and pets. &
Nemaha, Richardson, Pawnee, Montgomery, Page, and Fremont counties in Nebraska and Iowa: Stay cool and hydrated as heat index values are expected to reach 109 today and 106 tomorrow. Find shade and limit outdoor activities, especially during peak hours. Take a cool shower or visit a local cooling center if available. &
A heat index of 109 is expected in your area for today, and it's set to drop to 106 on Tuesday. It is crucial to stay cool and hydrated. Consider taking cool showers or baths if you feel overheated. Remember to check on elderly neighbors and family members, ensuring they are safe. \\
\midrule
-A bomb attack occurs at the tax office in Aix-en-Provence, France. &
**URGENT - Aix-en-Provence, France**

A bomb attack has occurred at the tax office on the Cours Mirabeau in Aix-en-Provence. {\color{red}Reports indicate damage and possible injuries.} Remain calm and follow instructions from local officials. Leave the area as quickly and safely as possible, without stopping to retrieve personal belongings. Follow the safest evacuation route and take all necessary precautions. &
**URGENT: Situation at Aix-en-Provence tax office**

A bomb attack has occurred at the tax office in Aix-en-Provence, France. **Please move to a safe location as quickly and calmly as possible. Follow local authorities' instructions and stay away from the area. {\color{red}Avoid any potential hazards or falling debris. If you are not directly involved,} stay informed through official updates and follow evacuation instructions.** &
**Warning**: An incident has taken place at the tax office in Aix-en-Provence, France. For your safety, stay away from the area and follow local officials' instructions. If you are near, move calmly to safety and do not stop to gather belongings. Remain informed through local news and updates. \\
\midrule
-Ohio River at Paducah currently experiencing widespread minor flooding.

-At 9:00 AM, the water level recorded at 40.3 feet, above the flood stage of 39.0 feet.

-Minor flooding is currently occurring and is anticipated to continue.

-The Flood Warning is in effect until Monday afternoon. &
**Important Notice for Paducah Residents:**

**Location:** Ohio River at Paducah
**Current Situation:** Minor flooding expected to continue throughout the day.

If you're in the affected areas, please prepare emergency supplies and stay informed about the latest updates from local authorities. Pay attention to higher ground and avoid walking or driving through floodwaters, as they can be contaminated. Follow all evacuation instructions and wait for authorities to confirm it's safe to return home. &
Attention citizens of Paducah. {\color{red}The Ohio River's water level is at 40.3 feet, above the flood stage.} Minor flooding is occurring, and the situation is expected to persist. Please stay indoors and be cautious of localized flooding. Avoid driving unless absolutely necessary, and follow local updates for further guidance. &
Flood Warning Issued: A Flood Warning remains active for the Ohio River at Paducah until Monday afternoon. Please stay tuned to authorities and avoid driving unless necessary. Always prioritize your safety when encountering floodwaters. \\
\bottomrule
\end{tabularx}
\caption{Comparison of the best setup for SFT, Zero-shot, and Few-shot. Text highlighted in red indicates example of the problems discussed in the Appendix \ref{app:zero-shot-sft-analysis}.}
\label{tab:zero-vs-few-vs-sft-comparison}
\end{table*}

\end{document}